\documentclass[11pt,a4paper]{article}
\usepackage{hyperref}
\usepackage[hyperref]{acl2020}
\usepackage{times}
\usepackage{latexsym}
\usepackage{verbatim}
\usepackage{xspace}
\usepackage{booktabs}
\usepackage{graphicx}
\usepackage{amssymb}
\usepackage{multirow}
\usepackage{xcolor}

\usepackage{adjustbox}
\usepackage{array}

\usepackage{etoolbox}
\makeatletter
\patchcmd\@combinedblfloats{\box\@outputbox}{\unvbox\@outputbox}{}{%
}%
 \makeatother

\newcolumntype{R}[2]{%
    >{\adjustbox{angle=#1,lap=\width-(#2)}\bgroup}%
    l%
    <{\egroup}
}

\usepackage{url}

\aclfinalcopy

\title{Can You Put it All Together:\\
Evaluating Conversational Agents' Ability to Blend Skills
}

\author{
  Eric Michael Smith*, Mary Williamson*, Kurt Shuster, Jason Weston, Y-Lan Boureau \\
  Facebook AI Research \\
  {\tt \{ems,marywilliamson,kshuster,jase,ylan\}@fb.com}
}

\date{}

\begin{document}
\maketitle

\begin{abstract}
Being engaging, knowledgeable, and empathetic are all
desirable general qualities in a conversational agent.
Previous work has introduced tasks and datasets 
that aim to help agents to learn those qualities
in isolation and gauge how well they can express them.
But rather than being specialized in one single 
quality, a good open-domain conversational agent
should be able to seamlessly blend them all into
one cohesive conversational flow.
In this work, we investigate several ways to combine
models trained towards isolated capabilities,
ranging from simple model aggregation schemes that require minimal additional training, to various forms of multi-task training that encompass
several skills at all training stages.
We further propose a new dataset, BlendedSkillTalk, to analyze how these capabilities would
mesh together in a natural conversation, and compare
the performance of different architectures and
training schemes. Our experiments show
that multi-tasking over several tasks that focus on particular capabilities results in better blended
conversation performance compared to models
trained on a single skill, and that
both unified or two-stage approaches perform well 
if they are constructed to avoid unwanted bias in skill selection
or are fine-tuned on our new task.
\end{abstract}

\section{Introduction}

A good open-domain conversational agent should have a well-rounded set of skills\footnote{"Skills" in the conversational AI literature is sometimes taken to mean
a very defined specific set of abilities such as telling the weather (e.g., \citet{zhou2018design}). Our use in this paper is much more general and refers to any desirable capability.} and qualities that allow it to seamlessly blend listening with empathy, providing knowledgeable responses, and talking about various topics from everyday life to their favorite hobbies or latest challenges.

Recent research has made solid strides towards gauging and improving performance of open-domain conversational agents along specific axes such as
how knowledgeable they are \citep{dinan2018wizard,moghe2018towards,qin2019conversing}, how well they can display empathy \citep{rashkin2019empathy,lin2019moel} or talk about their personal background \citep{zhang2018personalizing,dailydialogue}. 
However it remains unclear whether models optimized for performance along one of these axes can retain the learned skill while blending it with other desirable skills, or how to best conduct simultaneous training of multiple skills.

In this work, we compare several ways to combine tasks designed to evaluate and improve a single conversational skill, ranging from multi-task training over several datasets to training a top-level classifier to play the role of a dialogue manager and query the most appropriate single-skill pretrained model for a response.
In order to evaluate those methods, we propose a new English-language dataset, BlendedSkillTalk,  that blends several skills into a single conversation, and use it to evaluate methods with both automated metrics and human crowdsourced ratings across different axes. 

Our experiments show that existing single-skill tasks can 
effectively be combined to obtain a model
that blends all skills into a single conversational agent
if care is taken to make the dialogue agent
avoid unwanted biases when selecting the skill, or if
fine-tuning on blended data, or both. 
We propose methods that compare those competing approaches, and provide a detailed analysis
of their successes and failures.

\section{Related work}
While most commercial dialogue systems rely on hand-coded
narrow skills (e.g., see \citet{zhou2018design,ram2018conversational}), typically focusing on separate task-oriented features such as alarm setting, calendar entries, etc.,  we are interested in models that display various qualities in
open-domain dialogue. Further, we focus on  skills that can be learned end-to-end, as 
end-to-end learning affords the promise of better generalization to unseen domains.

Recent promising conversational models have leveraged very
large conversation-like data such as datasets extracted from Reddit and made available by a third party on pushshift.io \citep{mazare2018training,humeau2019real,keskar2019ctrl,rashkin2019empathy}. These large-scale datasets are very useful in providing vast
amounts of conversational material that allow for reproducible research and comparison with prior work, however 
the qualities of resulting conversational agents are dependent
on the qualities present in the source conversations. Given
how online conversations can turn toxic and lack empathy,
indiscriminate pretraining on such corpora is unlikely
to spontaneously endow a conversational agent with desirable
qualities such as avoiding toxic responses \citep{dinan2019build}
or demonstrating empathy \citep{rashkin2019empathy} or knowledge \citep{dinan2018wizard}.

This has led the community to propose tasks and datasets
focusing specifically on some trait or skill. In this work,
we examine how to combine three such traits that each have
a corresponding task and dataset: demonstrating
an ability to talk about oneself and get to know your partner,
as captured by the ConvAI2
dataset, an extension of the PersonaChat dataset \citep{zhang2018personalizing,dinan2019second}; 
being knowledgeable and discussing a topic in depth, as
measured through the Wizard of Wikipedia task \citep{dinan2018wizard}; and demonstrating empathy
and being able to talk about emotional personal situations, as measured by the EmpatheticDialogues benchmark proposed in \citet{rashkin2019empathy}.
The ConvAI2 dataset comprises more than 140k utterances of crowdsourced conversations between paired workers getting to know each other. Each worker was assigned a persona
consisting of a few sentences such as ``I have a pet hamster,'' which had separately been
crowdsourced. The Wizard of Wikipedia (WoW) task aims to explore conversation informed
by expert knowledge from Wikipedia, and provides about 194k utterances of conversations on about 1,250 topics. The EmpatheticDialogues (ED) dataset consists in about 50k utterances
between a Speaker who is talking about an emotional situation, and a Listener who is
tasked to respond in an empathetic manner, acknowledging the other person's feelings. In addition to being associated with easy-to-use datasets, these three skills benefit from being clearly defined and separate in scope. Focusing on blending only three skills keeps data collection, ablations, and analyses manageable while already presenting a challenge for models, and it helps narrow down the most promising approaches for blending a greater number of skills.


\section{Blending Skills in a Conversation}
A model separately trained on a variety of skills might be able to do well on each
of them in isolation, but still struggle to seamlessly blend them over the course 
of a single conversation where it has to navigate whether a given utterance
calls for informative knowledge or empathy, for example. It must learn to switch between skills, each time incorporating previous dialogue context which may contain utterances from either partner relating to multiple skills, and on some turns may have to blend skills into a single response.

\subsection{BlendedSkillTalk}\label{sec:bst}


In order to gauge how successful a model is at this blended objective, 
we collect BlendedSkillTalk, a
small crowdsourced dataset of about 5k conversations in English where workers are instructed
to try and be knowledgeable, empathetic, or give personal details about their given persona,
whenever appropriate. We collect conversations from 2,679 workers, with each worker participating in an average of 5.4 conversations in the train set and a maximum of 15 conversations. The dataset consists of 4,819 train-set conversations, 1,009 validation-set conversations, and 980 test-set conversations. We ensure that the sets of workers involved in collecting the train, validation, and test sets are completely disjoint to prevent our models from benefiting from learning about specific workers' biases \citep{geva2019we}. On average, there are 11.2 utterances (5.6 pairs from the two workers) in each conversation in the train set. This dataset is available through the ParlAI framework\footnote{\tt{https://parl.ai/}}.

 An example conversation from BlendedSkillTalk is shown in Figure \ref{tab:example_conversation}.
 In this example, we see that the speakers
 inject knowledge, empathy, and personal background, and generally that
 the conversation invokes different skills while flowing naturally.

\begin{figure*}[t!]
 \small
    \centering
    \begin{tabular}{l|l} 
    \toprule
    \textbf{Persona for \textcolor{red}{Unguided Speaker}}: & \textbf{Persona for \textcolor{blue}{Guided Speaker}}:\\
    My son plays on the local football team. & My eyes are green. \\
    I design video games for a living.       & I wear glasses that are cateye. \\
\midrule
\multicolumn{2}{l}{\textbf{Wizard of Wikipedia topic}: Video game design} \\
\multicolumn{2}{l}{\textbf{Previous utterances (shown to speakers)}:} \\
\multicolumn{2}{l}{\textbf{\textcolor{red}{U}}: What video games do you like to play?} \\
\multicolumn{2}{l}{\textbf{\textcolor{blue}{G}}: all kinds, action, adventure, shooter, platformer, rpg, etc. but video game design requires both artistic and technical } \\
\multicolumn{2}{l}{~~~~~competence AND writing skills. that is one part many people forget} \\
\midrule
\multicolumn{2}{l}{\textbf{Actual utterances}: }\\
\multicolumn{2}{l}{\textbf{\textcolor{red}{U}}: Exactly! I think many people fail to notice how beautiful the art of video games can be. \textbf{\textcolor{violet}{(PB)}}} \\
\multicolumn{2}{l}{\textit{\textcolor{gray}{(G selected the WoW suggestion: "Indeed, Some games games are purposely designed to be a work of a persons creative}}}\\
\multicolumn{2}{l}{\textit{\textcolor{gray}{expression, many though have been challenged as works of art by some critics.")}}} \\
\multicolumn{2}{l}{\textbf{\textcolor{blue}{G}}: Indeed, Some games games are purposely designed to be a work of a persons creative expression, many though have been}\\
\multicolumn{2}{l}{~~~~~challenged as works of art by some critics. \textbf{\textcolor{violet}{(K)}}} \\
\multicolumn{2}{l}{\textbf{\textcolor{red}{U}}: Video games are undervalued by many and too easily blamed for problems like obesity or violence in kids \textbf{\textcolor{violet}{(K)}} }\\
\multicolumn{2}{l}{\textbf{\textcolor{blue}{G}}: Indeed, Just last week my son was playing some Tine 2 and it was keeping him so calm. } \\
\multicolumn{2}{l}{~~~~~Games are therapeutic to some. \textbf{\textcolor{violet}{(S)}} } \\
\multicolumn{2}{l}{\textbf{\textcolor{red}{U}}: I use games to relax after a stressful day, the small escape is relaxing. \textbf{\textcolor{violet}{(PB)}}} \\
\multicolumn{2}{l}{\textit{\textcolor{gray}{(G selected the ED suggestion: "I enjoy doing that after a hard day at work as well.  I hope it relaxes you!")}}} \\
\multicolumn{2}{l}{\textbf{\textcolor{blue}{G}}: I enjoy a good gaming session after a hard day at work as well. \textbf{\textcolor{violet}{(PB)}}} \\
\multicolumn{2}{l}{\textbf{\textcolor{red}{U}}: What other hobbies does your son have? \textbf{\textcolor{violet}{(PB)}} }\\
\multicolumn{2}{l}{\textbf{\textcolor{blue}{G}}: Well he likes to fly kites and collect bugs, typical hobbies for an 8 year old, lol. \textbf{\textcolor{violet}{(PB)}} }\\
\multicolumn{2}{l}{\textbf{\textcolor{red}{U}}: My 12 year old is into sports. Football mostly. I however don;t enjoy watching him play. \textbf{\textcolor{violet}{(PB)}} }\\
\multicolumn{2}{l}{\textbf{\textcolor{blue}{G}}: I wish I could play football, But I wear this cateye glasses and they would break if I tried. \textbf{\textcolor{violet}{(PB)}}}  \\
\multicolumn{2}{l}{\textbf{\textcolor{red}{U}}: Sounds nice. Are they new or vintage? \textbf{\textcolor{violet}{(E)}}} \\
\multicolumn{2}{l}{\textbf{\textcolor{blue}{G}}: They are new, I got them because of my love for cats lol. I have to show off my beautiful green eyes somehow. \textbf{\textcolor{violet}{(S)}} }\\
\bottomrule  
\end{tabular}
    \caption{\label{tab:example_conversation} Sample conversation from the BlendedSkillTalk dataset, annotated with four conversation mode types 
    (PB: personal background; K: knowledge; S: personal situation; E: empathy).
    The guided (G) and unguided (U) workers are given personas and a topic. The conversation has been seeded with two utterances from a conversation sampled from WoW. When the guided worker selected one of the suggestions, it is shown in shaded grey.
    }
\end{figure*}

\paragraph{Guided Collection}
In order to prevent workers from getting stuck in a set ``mode'' of conversation (in which they consistently use one specific skill) or from being too generic, we provide responses
from models that have been trained towards a specific skill as inspiration to one of the two workers in the conversation. That worker is free to either use and modify or ignore those responses. Thus, each conversation involves an ``unguided" speaker and a ``guided" speaker, with the unguided speaker talking first. Whenever it is the guided speaker's turn to respond, we show them three suggested responses, one each from three single-task poly-encoder \citep{humeau2019real} models trained on the ConvAI2, ED, and WoW datasets. These are the same models we use as
baseline conversational agents for individual skills as well.

A breakdown of the choices of guided speakers is shown in Table~\ref{fig:statsGTctxt}, showing a reasonably balanced choice of suggestions. 
Workers decide to use them in 20.5\% of utterances, which affects the overall dialogues.
Interestingly, 46.1\%  of the time (versus 33.3\% at chance), the unguided speaker continues in the same mode as the previous utterance by the guided speaker, according to the classifier.
Thus, the BlendedSkillTalk dataset mimics natural conversation by featuring both continuity (``stickiness'' in the conversation mode) and mode blending within a single conversation.

\paragraph{Blended Initial Contexts}
Each speaker is assigned a pair of sentences from randomly-chosen personas from the ConvAI2 dataset. Similar to the ConvAI2 setting, each speaker sees their own persona but not that of the other speaker. Each conversation is seeded with a randomly selected pair of utterances from ConvAI2, WoW, or ED, with equal probability. Workers are instructed to continue the conversation from there. Workers are also provided with the topic being discussed if the conversation seed is from WoW, or the situation description if it is from ED. Note that this latter set-up departs from the ED  benchmark set-up, where the situation description is not used. The rationale for this is to provide some context about what was being discussed if the seed utterance pair happened to be extracted from the middle of a conversation. When WoW is used as seed, the chosen personas and the initial conversation topic are selected to match, similar to the original WoW paper.

To gain more insight into the influence of the datasets that provide this context, we leverage an utterance classifier trained to assign utterances to one of the three datasets (ConvAI2, WoW, ED; described further in Section~\ref{sec:arch}). We find that the average percentage of utterances from the unguided worker that match the provided context dataset is 43.5\% over the training set, compared to 33.3\% if the source of the provided context had no influence (note that this observed "stickiness" is similar to the 46.1\% of times the unguided speaker continues in the same mode as the one initiated by the guided speaker, mentioned above).
This suggests that the choice of seeding utterances and context indeed has an influence on the
type of blend observed, helping to make the dataset balanced. Table~\ref{tab:uttClassif} breaks down the classification results
by provenance of the seed context. The fraction of utterances resembling a given dataset
increases when the seed context is from that same dataset.
However the conversations are still blended: when breaking down the training set conversations
according to the number of ``modes'' observed in the utterances of the unguided worker according to the classifier,
47.8\% show 3 modes, 43.2\% show two modes, and 9.1\% show a single mode.

\begin{table}[t]
    \centering
    \begin{small}
    \begin{tabular}{llll}
       Chosen suggestion &  Initial Context & Count & Total  \\
    \hline
      \multirow{3}{*}{{\em none}} &  ConvAI2  &  7280 & \multirow{3}{*}{21468} \\ 
                                  &   ED     & 7257 \\
                                  &  WoW    &   6931 \\
      \midrule
      \multirow{3}{*}{ConvAI2}    &  ConvAI2 &  567 & \multirow{3}{*}{1599} \\  
       &  ED &  496 \\
       &  WoW &   536\\
        \midrule
      \multirow{3}{*}{ED}  &    ConvAI2 & 766 & \multirow{3}{*}{2221} \\ 
        &     ED &    773 \\
       &     WoW &     682 \\
      \midrule
      \multirow{3}{*}{WoW}     &    ConvAI2 &   634 & \multirow{3}{*}{1730} \\ 
       &    ED &    494 \\
       &    WoW &   602 \\
          \end{tabular}
          \end{small}
    \caption{
    Guided workers choice of suggestions in the train set of BlendedSkillTalk, broken down by provenance of the given initial context utterances. 
    Guided workers often choose not to use the suggestions, but have a slight preference for  
      ConvAI2 when the initial context is from that dataset, and similarly for ED.
   }
   \label{fig:statsGTctxt}
\end{table}

\begin{table}[t]
    \centering
    \begin{small}
    \begin{tabular}{p{2cm}rrr}
\toprule
& \multicolumn{3}{c}{Source of Seed Context}     \\
\cmidrule(lr){2-4}
\% classified as: & ConvAI2 & WoW & ED   \\
\midrule
ConvAI2 & \bf{29.6} & 25.3 & 25.5  \\
WoW & 49.6 & \bf{57.5} & 30.3  \\
ED & 20.8 & 17.1 & \bf{44.2}  \\
\bottomrule    
\end{tabular}
    \end{small}
    \caption{\label{tab:uttClassif}
Percentages of utterances of unguided workers classified by the dataset classifier
as coming from ConvAI2, WoW, or ED, broken down by provenance of the provided seed
context. For each dataset, the fraction of utterances classified as coming from that dataset
is highest when the seed context is from that same dataset.
    }
\end{table}

\paragraph{Data Quality}
To improve the quality of the collected conversations, we filter out any conversations where one of the speakers speaks less than 3 words per message; starts their conversation with a greeting despite previous utterances existing in the conversation; uses all-caps too frequently; repeats themselves too much; writes a message that gets flagged by a safety classifier; or, if they are the guided speaker, always accepts suggestions verbatim without changing them. Messages cannot be over 30 words or copy persona strings exactly.

\paragraph{Skill Annotations}
We also asked crowdsource workers to rate individual utterances as exhibiting one of four possible modes:
\begin{itemize}
    \item Knowledge: using factual information (\textit{``I’ve heard that in some places, lifeguards also help with other sorts of emergencies, like mountain rescues!"}) \citep{dinan2018wizard}
    \item Empathy: understanding and acknowledging implied feelings (\textit{``I’m sorry to hear that. I wish I could help you figure it out"}) \citep{rashkin2019empathy}
    \item Personal situations: past circumstances in a person's life (\textit{``I finally got that promotion at work! I have tried so hard for so long to get it!”}) \citep{rashkin2019empathy}
    \item Personal background: a person's personality, interests, and attributes (\textit{``I am into equestrian sports."}) \citep{zhang2018personalizing}
\end{itemize}
All utterances in over 700 conversations from the validation set of the BST dataset, from both guided and unguided workers, were annotated in this manner for 7,380 annotations collected in total. Workers were able to select as many attributes as they wished for each utterance. To avoid worker-specific bias, each crowdsource worker was limited to performing annotations on 10 conversations, and 123 total workers contributed annotations. Most analysis in this paper refers to three datasets, and the utterance classifier was trained with three dataset labels as classes. However, the ED dataset contains both ``Speaker'' utterances that describe personal situations, and "Listener" utterances, where the Listener responds with empathy (the ED benchmarks trains on both sides but evaluates only on the Listener side).
We therefore break down annotations into four types, with two types covering responses about ``personal topics": personal background (which is the focus of ConvAI2) and personal situations (talked about in ED). Results in Table~\ref{tab:annotationsPerConvo} show that the dataset indeed contains a reasonably balanced blend of these qualities. Over 70\% of conversations annotated contained at least 3 of 4 modes.
Overall, workers' annotation counts are 43.7\% for personal background, 20.5\% for knowledge, 20.3\% for empathy, and 15.4\% for personal situations. This supports the finding from our utterance classifier that the vast majority of conversations feature more than one mode, where utterance modes are defined as the predicted dataset provenance per utterance. In order to avoid excessive annotator bias and keep annotations discriminative, we limit the maximum number of annotations per worker and check that annotators did not select all modes for each utterance.

\begin{table}
    \centering
    \begin{small}
    \begin{tabular}{ccc}
\toprule
Mode Count & Conversations & Pct (\%)  \\
\midrule
1 & 51 & 6.9\%  \\
2 & 167 & 22.6\%  \\
3 & 290 & 39.2\%  \\
4 & 232 & 31.4\%  \\
\bottomrule    
\end{tabular}
    \end{small}
    \caption{\label{tab:annotationsPerConvo}
Breakdown of conversations by number of modes, showing that most BST dataset conversations exhibit multiple modes. Workers were asked to choose if each utterance of a conversation demonstrated knowledge, empathy, personal situations, or personal background. Over 70\% of the conversations annotated demonstrated at least 3 of the 4 modes.
    }
\end{table}

\subsection{Blending Skills in a Single Model}
\label{sec:arch}
\paragraph{Architectures and Training} The base architecture used throughout the paper is the 256-million parameter 
poly-encoder proposed in \citet{humeau2019real}, 
which is a  Transformer-based architecture for retrieval that learns a small number of codes representing the input context, so that performing attention over retrieval 
candidates is tractable in real-time, and was shown to be state of the art on
several datasets.
The polyencoder is first pretrained on the pushshift.io Reddit dataset and then fine-tuned on individual
datasets. 
At test time, these models retrieve from the set of training utterances to output a response.

Swept hyperparameters include dropout fractions, learning-rate schedule, the number of poly-encoder codes used to represent the context, the output scaling factor, and the output reduction type (max across outputs vs. mean across outputs vs. first output only). Hyperparameters that were held constant included a training batch size of 512 and learning with Adamax; 12 encoder layers and an embedding size of 768; and label and text truncation lengths of 72 and 360. Note this model discards all casing information.  Models were trained until validation-set hits@1 failed to improve for 10 epochs.
All training is conducted in ParlAI \citep{miller2017parlai}.

Model selection during fine-tuning is performed by choosing the model that scores highest on hits@1 on the validation set. This architecture is then leveraged
in different ways to combine different skills in a single agent.

\paragraph{Fine-tuning on the BlendedSkillTalk Dataset}
The simplest setting is to directly fine-tune the base architecture on a dataset that 
exhibits the blended skills we are looking for. In this setting, we simply fine-tune
the poly-encoder pre-trained on pushshift.io Reddit on the BlendedSkillTalk dataset, 
following the procedure in \citet{humeau2019real}. This setting is referred to as ``BST'' thereafter (for BlendedSkillTalk).

Such blended multi-skill training is only possible if a resource like BlendedSkillTalk is available, which we only just collected. Thus, interesting questions unanswered by such training include: (i) can we learn a strongly performing multi-skilled model with only individual tasks and no access to blended data? (ii) would a model with both individual skill training and blended skill training be superior?

\paragraph{Multi-task Single-Skills}
A straight-forward approach given access to multiple single-skill tasks is to multi-task on all of them during the fine-tuning step.
Using the multi-task training framework in ParlAI, we again start from the
poly-encoder pre-trained on pushshift.io Reddit, and fine-tune it multi-tasking on
ConvAI2, WoW, and ED.
The architecture is thus the same as for the single-task models, and has the same number of parameters. 
We select the model with the highest macro-average hits@1 across all training tasks.

\paragraph{Mitigating Single-Skill bias}

The straight-forward way of multi-tasking over single skills is to sample training data from each task during updates. 
However, if individual skill contexts are too different from each other a multi-task model
will trivially separate the learning, rather than blending skills together. Then, if the bias
is different at evaluation time, it will select the skill to use poorly. In our case, ConvAI2 dialogues 
include a persona context, while WoW includes a topic.
This difference runs the risk of biasing the multi-task model into associating the mere presence
of a persona context to chat about personal background, and that of a discussion topic to discussions where more knowledge is displayed, which could lead to over-emphasizing responses
in the ConvAI2 style when tested on BlendedSkillTalk which contains personas.

We thus also experiment with a multi-task setting where the single skills are modified to always include a persona and a topic, as this is then balanced, and corresponds to the final evaluation using BlendedSkillTalk.
For every dialogue in each of the single-skill tasks, we thus prepend a persona and a topic to the first utterance if they are not already present.
The personas and topics are selected from the training sets of ConvAI2 and WoW respectively, where WoW topics already 
have an alignment to ConvAI2. For WoW, a persona is selected via this mapping. For ConvAI2, a topic is found with the inverse mapping. For ED, the maximum word overlap between the first utterance of the conversation and any training set persona is used to select the appropriate persona, and then a topic is found as before.

\paragraph{Multi-task Single-Skills + BlendedSkillTalk}
After training in a multi-task fashion on single skills, we can afterwards try to continue training with the BlendedSkillTalk resource, in an effort to
improve the model's ability to deal with blended data. We take the best model
previously trained, and tune it in this fashion.

\paragraph{Multi-task Two-Stage} 
Many single-skill models have been trained and released by researchers.
Harnessing those trained models could potentially
allow a conversational agent to jointly exhibit all skills, with minimal additional training. Instead, one trains a top-level `dialogue manager' which is a classifier with the dialogue context as input, that predicts which skill to use on each turn,
and then outputs the utterance produced by the corresponding trained model.
Specifically, we train a three-class classifier on top of 
BERT-base \citep{devlin2019bert} that assigns an utterance to the dataset it came from.
We remove duplicate utterances present in more than one of the datasets prior to training and upsample with replacement to create equal representation in the classifier's training set. We also remove context from the utterances including topics from Wizard of Wikipedia and personas from ConvAI2 before training this classifier and when performing evaluation to prevent the classifier from relying on these (cf. the bias mitigation mentioned above).

\section{Experiments}

In Section~\ref{sec:metrics}, we introduce the automated metrics and human evaluations that we use to measure and compare model performance. Section~\ref{sec:bias} discusses how adding personas and topic strings during multi-task training de-biases the selection of retrieval candidates from across our three skill-based tasks. Sections~\ref{sec:ss_benchmarks} and~\ref{sec:bst_benchmark} detail the performance of our models using automated metrics on single-skill and BlendedSkillTalk benchmarks, respectively, and Section~\ref{sec:human_eval} compares the performance of the models on human evaluation: in all three cases, models trained on all three skills generally outperform those trained on individual skills.

\subsection{Metrics used}
\label{sec:metrics}
We use both automated metrics and human evaluation.
For automated metrics, we report hits@1 on the test set (or validation set in the case of ConvAI2
as the test set is not publicly available), out of 20 candidates for ConvAI2, and 100 candidates for ED and WoW, following the original datasets. 
For human evaluation, we ask workers to chat with various models
and then rate the conversation along several axes:
\begin{itemize}
    \item Knowledge: How knowledgeable was your chat partner (from 1: not at all, to 5: very)?
    \item Empathy: Did the responses of your chat partner show understanding of your feelings (from 1: not at all, to 5: very much)? 
    \item Personal: How much did your chat partner talk about themselves (from 1: not at all, to 5: a lot)?
    \item Overall: Overall, how much would you like to have a long conversation with this conversation partner (from 1: not at all, to 5: a lot)?
\end{itemize}

Conversations and ratings are collected at least 100 times per model, from 234 crowdsource workers who produce a maximum of 10 of these conversations overall (across all model types).
Several methods are used to filter out low quality workers that are similar to the methods used in collection of the BlendedSkillTalk dataset collection. All work by a given worker is excluded if they give the same ratings across all conversations, give utterances deemed unsafe by a safety classifier \citep{dinan2019build}, utterances shorter than 3 words, use all-caps too frequently, or repeat themselves too much. Messages cannot be over 30 words or copy persona strings exactly.

\begin{table}
    \centering
    \begin{small}
    \begin{tabular}{lcccc}
\toprule
& \multicolumn{2}{c}{MT  Single-Skills} &  \multicolumn{2}{c}{MT S.-S.  + BST }    \\
\cmidrule(lr){2-3} \cmidrule(lr){4-5}
Utt. Selected & orig. & debiased & orig.  & debiased \\
\midrule
ConvAI2 & 64.4\% & 38.9\% & 61.1\% & 48.1\% \\
WoW     & 11.3\% & 29.4\% & 10.0\% & 21.3\% \\
ED      & 24.2\% & 31.6\% & 28.8\% & 30.5\% \\
\bottomrule  
\end{tabular}
    \end{small}
    \caption{\label{tab:debias} 
    Mitigating skill selection bias. Adding personas and topics during multi-task training (debias) results in the multi-task retrieval models 
    selecting utterances more evenly when tested on 
    BlendedSkillTalk compared to training on the original datasets (orig).
    }
\end{table}

\subsection{Mitigating multi-task skill selection bias} 
\label{sec:bias}

We first examine the issue of skill selection bias in multi-task models. As we are employing multi-task 
retrieval models that retrieve from the set of candidates across all 
skills, we can collect statistics on those selection choices (i.e., which datasets the chosen utterances originated from). 
Table~\ref{tab:debias} reports the percentage
of utterances derived from the three skills for our multi-task models (MT Single-Skills and MT Single-Skills + BST) when
evaluating on the BST test set.
When training on the original skill datasets, we observe heavy overuse
of the ConvAI2 utterances and underuse of WoW, likely because BST contains personas as input. 
Our bias mitigation approach described in Section 
\ref{sec:arch} causes a substantial shift for both models, making the use of the skills more equal. These results are then in line with the actual expected ratios in BST, as shown in Section \ref{sec:bst} 
(Skill Annotations). 
In the following experiments, we thus use the debiased versions.

\begin{table}[t!]
    \centering
    \begin{small}
  \begin{tabular}{lrrrr}
\toprule
& \multicolumn{3}{c}{Single-skill benchmarks}     \\
\cmidrule(lr){2-4}
Model & ConvAI2 & WoW & ED  & Avg. \\
\midrule
SOTA Reported & 87.3 & 87.4 & \bf{66.0} & {\em 80.2} \\
\midrule
ConvAI2 & \bf{89.4} & 78.4 & 42.6 & 70.1 \\
WoW & 57.3 & \bf{91.8} & 47.7 & 65.6 \\
ED & 63.3 & 81.0 & \bf{65.1} & 69.8 \\
\midrule
BST model & 78.5 & 84.1 & 52.0  & 71.5\\
Random-Skill & 71.0 & 83.9 & 52.0 & 69.0 \\
MT Two-Stage & 84.7 & 90.1 & 63.4 & 79.4 \\
MT Single-Skills & \bf{88.8} & \bf{92.8} & \bf{63.2} & {\bf 81.6}  \\
\midrule
& \multicolumn{3}{c}{Added-context benchmarks}     \\
\cmidrule(lr){2-4}

MT Single-Skills & 88.9 & 92.8 & 63.2 & 81.6 \\
\midrule
& \multicolumn{3}{c}{Mixed-candidates evaluation}     \\
\cmidrule(lr){2-4}
Single-task & 82.1 & 88.2 &  60.2 & \em{76.8} \\
MT Two-Stage & 77.2 & 86.6 &  59.0  & 74.3 \\
MT Single-Skills& \bf{85.2} & \bf{92.1} & \bf{61.1} & {\bf 79.5}  \\
\bottomrule  
\end{tabular}
    \end{small}
    \caption{\label{tab:singleTask}
Results on single-skill benchmarks.
Top: reported values published in the papers accompanying the
benchmarks, and the Poly-encoder paper.
ConvAI2, WoW, ED: models trained on the corresponding benchmark.
These models perform very well on the benchmark they were trained on, but not as
well on other benchmarks.
BST: The model fine-tuned on BST shows more balanced performance (i.e., none of the single-skill
benchmarks does better at all three skills), but it is noticeably lower than each specialized model.
Random-Skill: the performance of choosing a random single-skill per response
is comparable to the BST model, but slightly worse on ConvAI2.
MT Two-Stage: guiding the generation by an actual task classifier as opposed to random selection
increases performance on all skills.
MT Single-Skills: this model performs best among the blended skills architectures, and nearly
matches the single-skill model performance (and surpasses it in the WoW case).
Added-context benchmarks: when the benchmark contexts are augmented with a persona and topic as
described in section~\ref{sec:arch}, the evaluation results barely change.
Mixed-candidates evaluation: when the set of benchmark candidates is tripled by adding candidates from 
the other two benchmarks in equal proportion, the performance of the best respective single-task models suffers, while the MT Single-Skills model proves more resilient. Note that  Single-task averages in italics do not correspond to a single model, but an average over 3 models.
    }
\end{table}

\subsection{Results on Single-Skill Benchmarks}
\label{sec:ss_benchmarks}
Automated metrics results on the original benchmarks used to gauge competency at a single skill (ConvAI2, WoW, ED) reported in the literature are shown in Table~\ref{tab:singleTask} (first row).
Our poly-encoder models (rows 2--4) trained on single tasks match or exceed the metrics published with the 
corresponding benchmarks, 
except for ED, which is close. 
The single-skill models each perform the best on their respective original benchmark and not as well on other benchmarks, compared to the blended models. However, the performance of all blended models is more balanced, in the sense that none of the single-skill models does as well averaged over the three categories (except for the ED model doing a tiny bit better than the random-skill model). The model fine-tuned on BST shows balanced performance but fails to match the performance of the single-skill models on their original benchmarks. The performance of the Multi-Task Two-Stage model gains many points over that of simple random assignment of single-skill models (Random-Skill), and this Random-Skill model itself performs about as well as the BST-fine-tuned model on the ED and WoW benchmarks. The Multi-Task Single-Skills model performs best among the blended models, and nearly matches the performance of all single-skill models on all benchmarks (even surpassing it for the WoW benchmark).

The fact that the  Multi-Task Single-Skills model does not do exactly as well as the single-skill models when evaluated using only candidates from individual benchmarks matches the observations of other work \citep{raffel2019exploring}.
However, when evaluated with a set of mixed candidates from all single-skill tasks (where the set of candidates to choose from is tripled by included an equal number of candidates from the other two datasets), the multi-task model performs better than the individual models, suggesting that multi-task training results in increased resilience to having to deal with more varied distractor candidates.
We also include metrics for ``added-context'', when topics and personas are added (see Section~\ref{sec:bias}), as a sanity check, but they indeed barely change the numbers on single-skill benchmarks.

\begin{table}[t!]
    \centering
    \begin{small}
    \begin{tabular}{lrr}
\toprule
Model & BST, zero-shot & +BST, FT\\
\midrule
ConvAI2 & 76.8 & 81.7 \\
WoW & 67.5 & 79.4 \\
ED & 69.0 & 80.4 \\
\midrule
BST & - & 79.2\\
Random-Skill & 71.2 & - \\
MT Two-Stage  & 71.9 & - \\
MT Single-Skills & \bf{80.1} & \bf{83.8} \\
\bottomrule  
\end{tabular}
    \end{small}
    \caption{\label{tab:blended}
Test results on BlendedSkillTalk. BST, zero-shot: the models are tested directly
on the test set of BST without having been fine-tuned on the BST train set.
+BST, FT: models are fine-tuned on the BST train set, then tested on the BST test set. Multi-Task Single-Skills + BlendedSkillTalk performs best.
The Multi-Task Two-Stage model outperforms  two of the single-skill models, but the latter work well when combined with BlendedSkillTalk fine-tuning.
We hypothesize that  ConvAI2 alone performs well because it has been trained to use persona contexts, that are used throughout the BST dialogues.
    }
\end{table}

\begin{table*}[t!]
    \centering
    \begin{small}
    \begin{tabular}{lrrrr}
\toprule
Model &  Knowledge & Empathy & Personal   & Overall quality \\
\midrule
ConvAI2 & 3.2 & 3.1  & \bf{3.4}   & 3.0 \\
WoW & 3.3 & 2.9 & 2.7  & 2.6\\
ED & 3.4 & 3.3 & 3.0  & 3.0 \\
\midrule
BST & \bf{3.5} & \bf{3.6} & 3.1  & 3.3\\
Random-Skill & 3.2 & 2.9 & \bf{3.2}  & 2.7 \\
MT Two-Stage  & \bf{3.7} & \bf{3.6} & \bf{3.3}  & \bf{3.5}\\
MT Single-Skills & \bf{3.7} & \bf{3.6} & 3.0  & \bf{3.4}\\
MT Single-Skills +BST fine-tuning & \bf{3.7} & \bf{3.8} & \bf{3.2}  & \bf{3.6}\\
\bottomrule  
\end{tabular}
    \end{small}
    \caption{\label{tab:blendedHuman}
Human evaluation results on individual axes of knowledge, empathy, and being personal, as well as overall quality. All results here have a 95\% confidence interval of $\pm$ 0.2 or 0.3, omitted to avoid cluttering the table. Results that are within the confidence interval of the best model performance are bolded.
ConvAI2, WoW, ED: models pre-trained on pushshift.io Reddit and fine-tuned on the
respective datasets. For Empathy and Personal topics, the individual models tend to do better when trained on a dataset tailored for that, however they all perform similarly
on the Knowledge dimension.
BST: model pre-trained on pushshift.io Reddit and fine-tuned on BST. This model is showing better overall performance compared to single-skill datasets (i.e., none of the three single-skill dataset do better than BST in every dimension). 
MT Single-Skills with fine-tuning on BST and MT Two-Stage  are performing very well
on all dimensions. MT Single-Skills with fine-tuning on BST has fewer than a third of
the parameters of the MT Two-Stage  model, yet manages to perform as well, if not slightly better.
    }
\end{table*}

\subsection{Results on BlendedSkillTalk benchmark}
\label{sec:bst_benchmark}

We show two types of results on the BlendedSkillTalk benchmark (BST). Single-skill models are tested directly on BST without any additional training in a zero-shot setting, or fine-tuned on the BST training set then tested on the BST test-set. Results for both settings are shown in Table~\ref{tab:blended}. 
The Multi-Task Single-Skills model outperforms all single-skill model baselines, whether used in a zero-shot or fine-tuned fashion, despite being the same size. The MT Two-Stage and Random-Skill models outperform two of the three single-skill models. We hypothesize that the ConvAI2 model is doing better because it has already learned to use personas.
All single-skill models show improved performance once fine-tuned on the BST train set. However, performance in the zero-shot setting is already good, which is promising in terms of generalization to unseen data.

\subsection{Human Evaluation on Specific Skill Axes}
\label{sec:human_eval}

Human evaluation results are shown in Table~\ref{tab:blendedHuman}. Single-skill models tend to generally be rated better than the other single-skill models on the skill they were optimized for, although all single-skill models are similarly rated on the knowledge axis.
Models that have been trained on multiple skills, either through multi-tasking (MT Two-Stage or MT Single-Skills) or through fine-tuning on BST, are performing well on every dimension, with the MT Two-Stage model and the MT Single-Skills fine-tuned on BST being the overall best. These two models have different advantages: the MT Single-Skills model fine-tuned on BST is more compact, being the same size as each individual single-skill model, but requires joint multi-task training, then fine-tuning. The MT Two-Stage model only requires training a classifier to play the role of a dialogue manager by assigning utterances to one of the three single-skill benchmarks, but is overall a much bigger model, given that it uses large models for each single skill and  the classifier itself.
The "Random-Skill" model is bypassing the need for a classifier by simply using all three single-skill model randomly, and is rated well on the personal axis, but not as well on knowledge or empathy, which might be because talking about personal topics can always work, while knowledge and empathy have to be suited to the context. 

\section{Discussion and Conclusion}
This paper focuses on the goal of creating an open-domain conversational agent
that can display many skills, and blend them in a seamless and engaging way. 
We have shown several ways to leverage
previous work focusing on individual conversational skills,
either by combining trained single-skill models
in a two-stage way,
by re-using the datasets for simultaneous multi-task
training, and by fine-tuning on the overall blended task. 
We compared the performance of these schemes
on BlendedSkillTalk, a new English-language dataset blending
three conversation skills in balanced proportions
(demonstrating knowledge, empathy, or ability to talk
about oneself). 
We showed that
multiple multi-task approaches can be effective on this task, however
careful construction of the training scheme is important
to mitigate biases when blending and selecting skills, while fine-tuning
on the overall blended task improves models further.

One natural extension would be to generalize these findings to other skills than the three addressed here, such as humor/wit, eloquence, image commenting, etc. This would in principle be straightforward to do as long as these additional skills have a corresponding ``single-skill" dataset to train on and are sufficiently distinguishable from each other.

\bibliography{main}
\bibliographystyle{acl_natbib}
\end{document}